# SCALING LARGE VISION-LANGUAGE MODELS FOR ENHANCED MULTIMODAL COMPREHENSION IN BIOMEDICAL IMAGE ANALYSIS

*Robinson Umeike*[1,2*] *Neil Getty*[1] *Fangfang Xia*[1] *Rick Stevens*[1]

[1] Argonne National Laboratory, Lemont, USA
[2] University of Alabama, Tuscaloosa, USA.

## ABSTRACT

Large language models (LLMs) have demonstrated immense capabilities in understanding textual data and are increasingly being adopted to help researchers accelerate scientific discovery through knowledge extraction (information retrieval), knowledge distillation (summarizing key findings and methodologies into concise forms), and knowledge synthesis (aggregating information from multiple scientific sources to address complex queries, generate hypothesis and formulate experimental plans). However, scientific data often exists in both visual and textual modalities. Vision language models (VLMs) address this by incorporating a pretrained vision backbone for processing images and a cross-modal projector that adapts image tokens into the LLM dimensional space, thereby providing richer multimodal comprehension. Nevertheless, off-the-shelf VLMs show limited capabilities in handling domain-specific data and are prone to hallucinations. We developed intelligent assistants finetuned from LLaVA models to enhance multimodal understanding in low-dose radiation therapy (LDRT)—a benign approach used in the treatment of cancer-related illnesses. Using multilingual data from 42,673 articles, we devise complex reasoning and detailed description tasks for visual question answering (VQA) benchmarks. Our assistants, trained on 50,882 image-text pairs, demonstrate superior performance over base models as evaluated using LLM-as-a-judge approach, particularly in reducing hallucination and improving domain-specific comprehension.

## 1. INTRODUCTION

Large language models (LLMs) are rapidly transforming scientific research due to their ability to create plausible natural language outputs supporting tasks such as text generation, translation, question answering, and content summarization [1]. However, processing scientific data often requires comprehending modalities beyond text. Vision language models (VLMs) address this limitation by introducing visual data comprehension for richer information retrieval.

VLMs, trained on extensive visual and textual datasets, exhibit considerable potential as intelligent assistants. Their training process involves two key phases: adapting vision tokens from the foundation model into the LLM space (feature alignment) and enabling specific downstream tasks (fine-tuning). In this study, we fine-tune LLaVA v1.6-vicuna-13B and LLaVA v1.5-13B on machine-generated biomedical data derived from low-dose radiation therapy (LDRT) articles [2], specifically using captions and textual artifacts related to the extracted images. Through a series of comprehensive experiments on our curated dataset, we demonstrate substantial improvement in LDRT-based visual question answering tasks compared to base models.

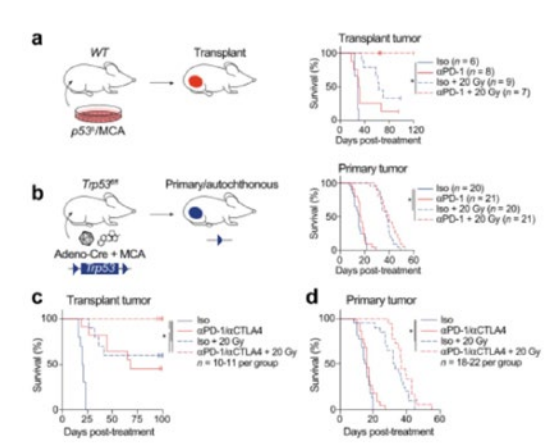

**Image Caption from Article**: Fig. 1 Immune checkpoint blockade and radiation therapy cure transplant but not primary sarcomas. a Transplant tumor initiation by p53/MCA cell injection into the gastrocnemius. Mice were treated with anti (α)-PD-1 (red) or isotype control (blue) antibody and 0 (solid) or 20 (dashed) Gy when tumors reached >70mm3. b Primary sarcoma initiation by intramuscular injection of Adeno-Cre and MCA. Treatment as in panel (a). c Mice with…

*Question*: What information can be understood from this figure?
***LLaVA v1.5***: The figure displays an overview of the complex process involved in transplanting a kidney into a patient. It consists of four main panels that show the various steps and treatments required. Some of the key findings include the donated kidney, grafting techniques, transplantation operation, and post-transplant treatment. The process involves removing the kidney from a donor (Panels A, B, and C), and then grafting it into the recipient (Panels D, E, and F). During this operation, several other factors, such as ensuring an immune-compatible match and managing possible complications, are crucial. After the transplant, the patient receives ongoing care and treatment with various medications, as represented by the line of medications in the figure. In addition, there are various other elements visible in the image that might have clinical purposes or are important steps in the process. Furthermore, there's a timeline representation of the immuno-competent recipient following transplantation and engraftment (Panel G). This figure illustrates the multifaceted nature of a successful kidney transplant operation from various perspectives.

*Question*: What information can be understood from this figure?
***LLaVA v1.6***: The figure is a series of graphs and illustrations that likely represent scientific data or results, possibly from a biomedical or cancer research study. It shows four panels (a, b, c, d) each with corresponding graphs and illustrations.
**A:** Tumor Burden (Weight [g] vs. Days post injection [DPI])
This panel shows the weight of tumors (burdens) over time, with different colors representing different treatment groups, such as "control," "p53/MCAb," "PD-1," and "PD-1 + 20 Gy." There is a mouse illustration with a tumorous lump, indicating where the tumor was implanted or grown.
**B:** Tumor Burden (Weight [g] vs. Days post treatment [DPT])
This panel continues the trend of showing tumor burden and includes a timeline with the beginning of treatment. The treatment groups are "CTLA-4," "CTLA-4 + 20 Gy," and "CTLA-4 + Tregs." Again, the graph shows the weight of the tumors at different times after treatment.
**C:** Survival (Percent vs. Days post treatment [DPT])
The third panel presents a survival curve, with the vertical axis showing the percentage of mice surviving, and the horizontal axis representing days after treatment. The treatment groups are "control," "PD-1," "CTLA-4," and "CTLA-4 + 20 Gy." This graph typically indicates the efficacy of treatments in prolonging life.
**D:** Survival (Percent vs. Days post treatment [DPT])
This is another survival curve, and it appears to show the same treatments as in panel C, but with additional data points suggesting a broader range of time for the post-treatment evaluation of survival.
The figures also contain mathematical formulas or treatment schemes, which likely describe the experimental design or the equations used to analyze the data. These formulas might be related to the calculations of the weights or survivals, or they could represent the specific treatment regimens used in the study.
Overall, the figure is informative and seems to document the effectiveness of various cancer treatments in reducing tumor weights and improving survival in cancer-bearing mice. The treatments range from chemotherapy (possibly represented by "20 Gy") to immunotherapy (represented by "PD-1," "CTLA-4," and other antibodies or agents). However, without the specifics of the research's methodology, these are general interpretations based on the visual elements and common practices within the field of cancer research.

**Fig. 1.** Qualitative examples showing hallucinations (highlighted in yellow) versus factually accurate statements (highlighted in green) for the base models [2, 10].

*Work was done when author was an intern at Argonne National Laboratory.
Corresponding author: Neil Getty (ngetty@anl.gov).

## 2. RELATED WORKS

**Language-Only Scientific Assistants**: Domain-specific LLMs have emerged as resourceful digital assistants or chatbots capable of engaging in meaningful dialogue across a variety of disciplines [3]. These models are typically trained on large, carefully curated corpora of finetuned textual data, enabling them to generate human-like responses that often surpass general-purpose models on field-specific benchmarks. For instance, Galactica, an open-source language model developed by Meta AI, demonstrates advanced capabilities in processing technical knowledge across multiple modalities including LaTeX, text, DNA/AA sequences, code, and SMILES notation [1]. Trained on over 48 million scientific documents, Galactica with 120B parameters outperforms general-purpose models like GPT-3 175B and PaLM 540B [3, 4]. Similarly, domain-specific models like SciBERT and BioBERT have achieved state-of-the-art performance in biomedical applications and benchmarks such as BC5CDR and NCBI-disease [5-8]. However, their effectiveness in tasks requiring visual comprehension remains limited due to the lack of multimodal understanding.

**Generalist VLMs**: Recent breakthroughs in general-purpose VLMs that process multiple modalities—images, videos, and audio alongside text—have sparked investigations into their adaptability for specialized applications [2]. These versatile models are gaining widespread recognition due to their potential relevance in many real-world tasks, such as analyzing medical imaging datasets for surgical planning or interpreting microscopy images for pathology diagnosis [9]. However, a major hurdle to adapting these generalist VLMs and LLMs to domain-specific benchmarks is the prevalence of hallucinations. In VLMs, this phenomenon occurs when the model makes up information that isn't in the source material (ground truth text or image), which presents a critical challenge in biomedical imaging applications where accurate interpretation of visual and linguistic data is crucial. For example, hallucination could lead to incorrect interpretations of X-ray or MRI scans, potentially compromising tasks such as automated diagnosis, report generation, and clinical decisions (Figure 1). While some studies have proposed increasing image resolution as an approach to mitigate this problem [10], our research takes a different approach. We seek to bridge this gap in biomedical image analysis by fine-tuning pretrained generalist VLMs specifically for scientific literature, with emphasis on preserving factual accuracy and reducing hallucinations through specialized preprocessing techniques and domain-specific training data.

## 3. METHOLODY

### 3.1. Data Preprocessing

GPT-4 was used to generate 100 keyword search strings related to LDRT, cancer, radiation biology, and related fields [3]. Semantic Scholar's API was then used to bulk download articles matching these keywords. After deduplication to obtain a set of unique open access paper IDs and abstracts, about 1 in 8 papers were successfully downloaded. In total, we acquired 42,673 LDRT-related scientific articles using Semantic Scholar. These articles were converted from PDF to Markdown format using Marker[1]. Then, we extracted images and captions using pdf2figures [11]. This process yielded nearly 171k captions and 165k images.

**Image-Caption Pair with Contextual data.** Following this, we identified low-resolution and blurry images using a Laplacian variance method to filter out images with a variance below 100. We set minimum dimensions of 100 pixels for both width and height to address low-resolution images. This process results in 150k images and caption pairs. From the cleaned captions, we extracted all instances where figures and tables were mentioned in the markdown files to provide better context for the images.

**LLM-generated data**. We prepare the system and user prompts for Qwen2-72B-Instruct to generate high-quality detailed description and complex reasoning question and answer pairs based on the captions and the contextual sentences [12]. For detailed descriptions, we defined 20 template questions (e.g., "Describe the contents of this image in detail", "What is this image about? ", "Analyze the figure in a comprehensive manner") and matched them to LLM-generated answers using term frequency-inverse document frequency vectorization and cosine similarity. Complex reasoning questions were specifically generated based on caption and context. This procedure resulted in 52456 entries split into training (50,882, 97%) and evaluation (1,574, 3%).

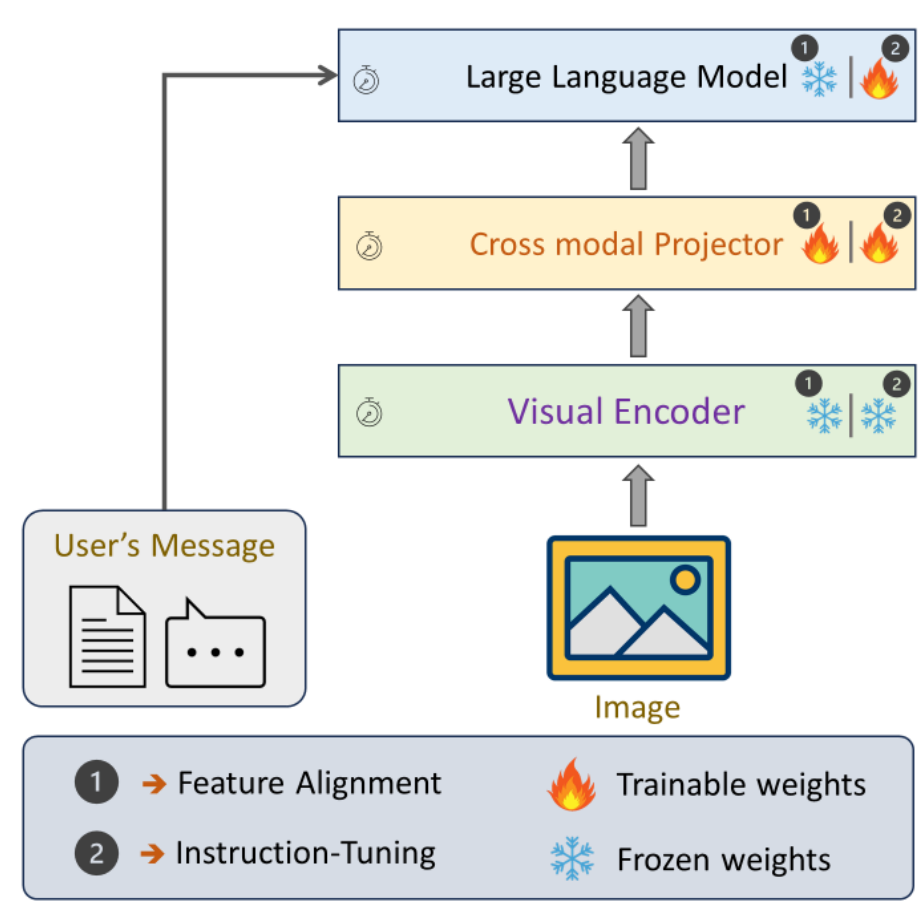

**Fig. 2.** Architecture overview showing projector alignment and instruction fine-tuning stages.

### 3.2. Model Architecture

Our model builds upon the LLaVA architecture (Figure 2), which consists of three main components: a pre-trained language model, a vision encoder, and a cross-modal projector [2]. The model's open-source nature and well-documented architecture make it ideal for research

[1] https://github.com/VikParuchuri/marker.

reproducibility and further adaptations. Let's formally define the model's components and their interactions. Given an input tuple (I, Q, A) representing an image, question, and answer respectively, the model processes these through the following pipeline:

**Visual Encoding.** $\psi$ represents the CLIP ViT-L/14 encoder, $\theta_v$ are the vision encoder parameters, $d_v$ is the vision embedding dimension, and n is the number of image patches.

$$V = \Psi(I;\ \theta_v) \in \mathbb{R}^{dvxn}$$

**Cross-Modal Projection**. $\Phi$ is a 2-layer MLP projector with parameters $\theta_p$ that maps visual features to the LLM embedding space of dimension $d_{llm}$.

$$V' = \Phi(V;\ \theta_p) \in \mathbb{R}^{dllmxn}$$

**Language Processing**. $\theta_l$ represents the Vicuna-13B parameters, and $a_t$ denotes tokens in the answer sequence at timestep t, $a_{<t}$ are the previously generated tokens before t.

$$P(A|Q,I) = \prod_t P(a_t|a <_t, Q, V';\ \theta_l)$$

The training process involves two distinct phases, which are **projector alignment** [$L_1(\theta_p) = \text{argmin}_{\theta_p} -\sum \log P(A|Q,I;\ \theta_p)$], for optimizing only the projector parameters $\theta_p$ while freezing $\theta_v$ and $\theta_l$; and **instruction finetuning** [$L_2(\theta_p, \theta_l) = \text{argmin}_{\theta_p,\theta_l} - \sum \log P(A|Q,I;\ \theta_p, \theta_I)$] for fine-tuning both projector and LLM parameters while keeping vision encoder frozen.

**Table 1**: Evaluation of base and fine-tuned models across detailed description and complex reasoning tasks.

| Judge | Model | Detail Mean ± Std | Complex Mean ± Std | Overall Mean ± Std |
|---|---|---|---|---|
| Qwen2-72B-Instruct [13] | LLaVA v1.5[2] | 1.74 ± 0.79 | 5.48 ± 2.37 | 3.46 ± 2.53 |
| | **ours** | **4.10 ± 2.90** | **6.63 ± 2.66** | **5.26 ± 3.06** |
| | LLaVA v1.6[10] | **2.66 ± 2.12** | 4.14 ± 2.28 | 3.34 ± 2.32 |
| | **ours** | 2.43 ± 2.28 | **6.52 ± 2.49** | **4.31 ± 3.14** |
| Llama3.1-70B-Instruct [19] | LLaVA v1.5[2] | 2.97 ± 1.19 | 6.24 ± 1.36 | 4.48 ± 2.07 |
| | **ours** | **5.97 ± 3.45** | **7.85 ± 1.94** | **6.69 ± 3.01** |
| | LLaVA v1.6[10] | **4.79 ± 2.07** | 6.55 ± 1.50 | 5.60 ± 2.03 |
| | **ours** | 3.86 ± 3.02 | **7.60 ± 1.65** | **5.60 ± 3.11** |

### 3.3. Finetuning and Optimization

Finetuning Large VLMs is a computationally intensive task requiring careful balance among model performance, runtime efficiency, GPU utilization, memory constraints, dataset size, and hyperparameter configurations. The optimization objective during fine-tuning can be formalized as:

$$\Theta^* = argmin_{\theta_p,\theta_I} [L_2(\theta_p, \theta_l) + \lambda R(\theta_p, \theta_I)]$$

where $L_2$ is the primary loss function, R is a regularization term, and $\lambda$ is the regularization coefficient controlling its contribution. To efficiently optimize this objective, we employ the following techniques:

**Memory Optimization**. We employed two key strategies to optimize memory usage: Gradient checkpointing, which reduces memory footprint by recomputing intermediate activations during backpropagation, and FlashAttention-2, which implements memory-efficient attention computation with O(N) (linear) memory complexity instead of the traditional $O(N^2)$ (quadratic) [14].

**Computational Efficiency**: DeepSpeed ZeRO3 (Zero Redundancy Optimizer) enables efficient model and data parallelism and portioning [15]. This partitioning strategy, combined with FlashAttention-2 [13], achieves significant runtime acceleration.

**Parameter Efficiency**. Low-Rank Adaptation (LoRA) was used to minimize the trainable parameter space during fine-tuning, $W = W_0 + BA$, where $W_0 \in \mathbb{R}^{dxk}$ represents the frozen pretrained weights, $B \in \mathbb{R}^{dxr}$, $A \in \mathbb{R}^{rxk}$ are trainable low-rank matrices, and r is the adaptation rank, such that $r \ll \min(d,k)$ [15]. Without these optimization strategies, training would be computationally infeasible given our available resources.

## 4. EXPERIMENT AND RESULT

Our training was initialized from the instruction-tuned LLaVA checkpoint. The experimentation was performed using a compute node equipped with 4 A40G GPU, training on 50,882 entries for 1 epoch with a batch size of 16 (4 per GPU × 4 GPUs) and a learning rate of 2e-4 with cosine decay. The model was configured with LoRA parameters (r=128, α=256) for both the vision module and the base model. The cross-modal projector, implemented as a 2-layer MLP with GELU activation, was trained with a learning rate of 2e-5. We utilized a 3% warmup ratio for learning rate scheduling and set the maximum sequence length to 2048 tokens.

**Model Evaluation and Inference.** For model evaluation, we performed inference on both base LLaVA models (v1.5-13B and v1.6-vicuna-13B) and our fine-tuned variants. The inference process utilized a temperature (balance between creativity and predictability) of 0.2 with a maximum generation length of 1024 tokens. Each image-question pair was processed by prepending image tokens to the question text, followed by model generation of responses. We evaluated models using both the base checkpoints from LLaVA and our fine-tuned versions, testing them on our evaluation set using identical image inputs and question prompts to ensure fair comparison.

**VQA Evaluation.** We used Qwen2-72B-Instruct [12] and Llama-3.1-70B-Instruct [16], as independent judges to evaluate both base and fine-tuned models on our image-text evaluation set. Following the evaluation protocol in LLaVA [2], the judges were prompted to assign integer scores (0-10) based on response relevance, helpfulness, and accuracy. The consistency in scoring between judges validates our evaluation methodology, as illustrated in Table 1.

**Hallucination Analysis.** The models' hallucination tendency was estimated using two complementary approaches. First, using ROUGE metrics to measure response alignment with ground truth (captions and contextual sentences), where higher scores indicate better factual consistency. Second, by analyzing linguistic markers of uncertainty, where we observed base models frequently employ hedging language. Notably, LLaVA v1.6-vicuna-13B uses "appears" 1,451 times in the evaluation set compared to only 49 times in our fine-tuned version. This significant reduction in hedging language, combined with improved ROUGE scores and length ratios closer to ground truth (Figure 4), suggests our fine-tuned models exhibit both higher confidence and better factual consistency.

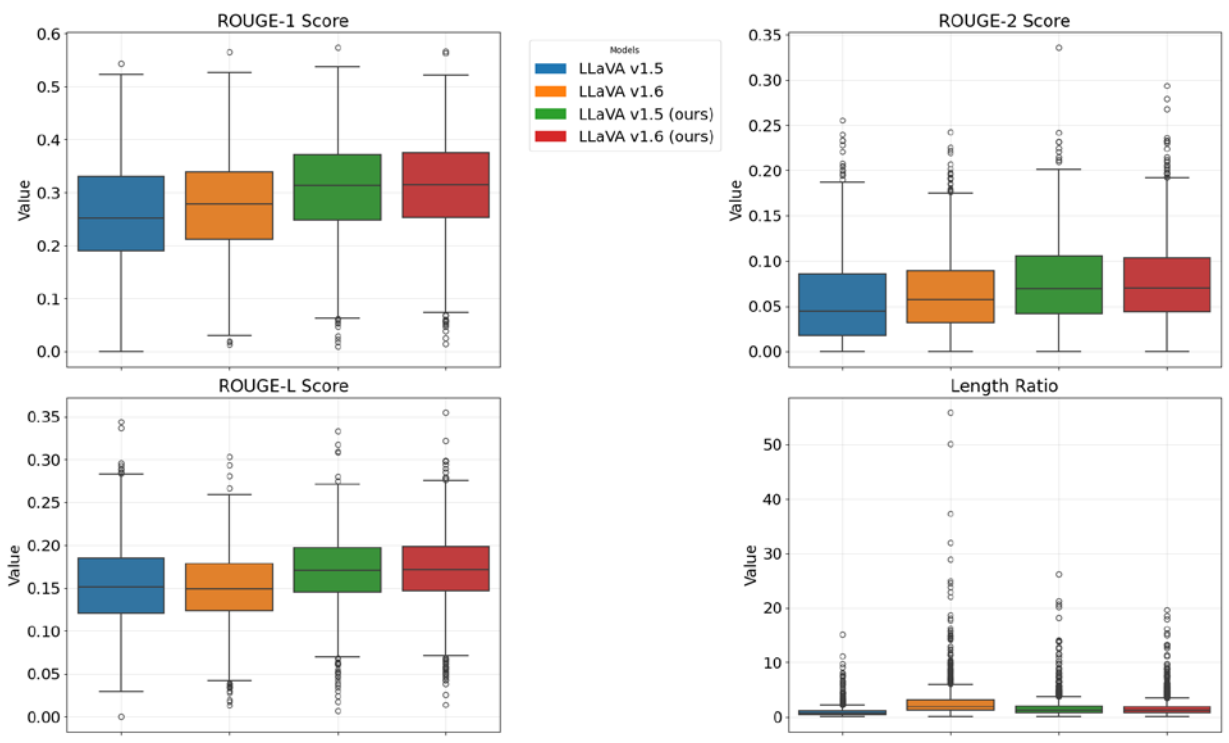

**Fig. 4.** Box plots show the distribution of scores across all 1574 evaluation set, with higher ROUGE scores indicating better alignment with ground truth and length ratios closer to 1 indicating more concise responses.

## 5. DISCUSSION AND CONCLUSION

In this paper, we present two scientific assistant models, fine-tuned from two LLaVA versions and capable of understanding LDRT images from scientific articles. They generally outperform LLaVA across model versions and question types; however, we observe an interesting trade-off in LLaVA v1.6-vicuna-13B, which shows better performance on detail-type tasks but exhibits verbosity as seen in its length ratio. This suggests a potential balance in our model between verbosity and reasoning depth. Given that complex reasoning represents a more challenging and valuable capability in scientific applications, our models' clear advantage in these tasks, combined with reduced hallucinations and improved confidence, is particularly significant. Future work will focus on expanding the data curation and hallucination analysis pipeline to enhance performance further and adapting the data for other specific biomedical applications.

## 6. ACKNOWLEDGMENTS

Argonne National Laboratory's work on the LUCID: Low-dose Understanding, Cellular Insights, and Molecular Discoveries program was supported by the U.S. DOE's Office of Science BER program, under Contract DE-AC02-06CH11357. The authors declare no competing interests.